# Satellite image data downlink scheduling problem with family attribute: Model &Algorithm

Zhongxiang Chang[1,2]※, Zhongbao Zhou[1,2]

[1]School of Business Administration, Hunan University, Changsha, China, 410082

[2]Hunan Key Laboratory of intelligent decision-making technology for emergency management, Changsha, China, 410082

## Abstract

The asynchronous development between the observation capability and the transition capability results in that an original image data (OID) formed by one-time observation cannot be completely transmitted in one transmit chance between the EOS and GS (named as a visible time window, VTW). It needs to segment the OID to several segmented image data (SID) and then transmits them in several VTWs, which enriches the extension of satellite image data downlink scheduling problem (SIDSP). We define the novel SIDSP as satellite image data downlink scheduling problem with family attribute (SIDSPWFA), in which some big OID is segmented by a fast segmentation operator first, and all SID and other no-segmented OID is transmitted in the second step. Two optimization objectives, the image data transmission failure rate (FR) and the segmentation times (ST), are then designed to formalize SIDSPWFA as a bi-objective discrete optimization model. Furthermore, a bi-stage differential evolutionary algorithm(DE+NSGA-II) is developed holding several bi-stage operators. Extensive simulation instances show the efficiency of models, strategies, algorithms and operators is analyzed in detail.

## 1. Introduction

With technology development of the platform and the payloads installed in earth observation satellites (EOSs), the observation capability of EOSs is improved significantly. On the one hand, the new generation EOSs can observe more ground targets for their three degrees of freedom (pith, roll, and yaw) (Lemaître et al., 2002) and more flexible attitude maneuver, named as active imaging (Chang et al., 2019, Lu et al., 2021). On the other hand, the resolution of EOSs has been increasing significantly in China. It is from "10 m resolution (Wang et al., 2000)" to "1 m resolution (Tong et al., 2016)" and improved again to "submeter resolution (Huang et al., 2018)" and even reaches "ultra-high resolution (Jones, 2018)", which results in the volume of each single scene image (one observation) increase exponentially. In addition, the constellation (Sai et al., 2018) is another main trend.

Since the development of observation capability, EOSs can capture more geographic information with much higher resolution. However, the technology for transmitting image data

---

※ Corresponding author: Zhongxiang Chang, E-mail address: zx_chang@163.com or zx_chang@hnu.edu.cn

is relatively backward, which makes the process of satellite image data downlink become more critical for capturing geographic information. For example, the constellation of Super View (Sai et al., 2018) is the first commercial satellite constellation with agile and multi-mode observation capability in China. Currently, the four on-orbit EOSs all have submeter resolution, but the speed for transmitting image data only reaches 2×450Mbps. An original image data (OID) generated by one observation of some long strips cannot be transmitted completely in a visible time window (VTW), which results it is imperative that segmenting an OID into several fragments, named as segmented image data (SID), and then transmitting them in serval different VTWs. Note that, how to segment OID is the first focus in our study. On the other hand, the quantity of ground stations is very scarce and most of them are located in the mainland of China (Guo et al., 2012), the distribution of them will described in the section 5.1.1. Since the scarce ground stations, optimizing the scheme for transmitting OID become more difficult, which is the second focus in this paper.

With the serial scheduling scheme in mind, the orthodox SIDSP can be viewed as a problem of optimizing a permutation of the downlink requests (Karapetyan et al., 2015). SIDSP and its variations have been studied by many authors. Some of these works focused on a single satellite (Karapetyan et al., 2015, Peng et al., 2017, Chang et al., 2022) whereas others were more general purpose in the special satellite constellation or multi-satellite (Bianchessi et al., 2007, Wang et al., 2011, Al et al., 2004, Chang et al., 2020, Chang et al., 2021a, Chang et al., 2021b). On the other hand, some researchers saw SIDSP as a time-dependent or resource-dependent problem and focused on the time-switch constraints between EOSs and ground stations (Du et al., 2019, Marinelli et al., 2011, Verfaillie et al., 2010, Zhang et al., 2019, Zufferey et al., 2008). In addition, other authors (Hao et al., 2016, Li et al., 2014, Peng et al., 2017, Wang et al., 2011) saw observation as an uncertain process, they considered observation scheduling problem for EOS (OSPFEOS) and SIDSP together or transformed SIDSP as constraints for OSPFEOS. But there is the same assumption in all proposed researches, that is, once a downlink starts, it cannot be preempted, i.e. an OID cannot be split into several fragments. Based on this hypothesis, SIDSP like a one-dimensional bin packing problem (BPP) with a time window in essence, the process of which is shown in Figure 1. Each OID likes an item will be packed into potential bins, the visible time windows (VTWs) between EOSs and ground stations as the potential bins. SIDSP just needs to select suitable OID to pack into VTWs, and then form the downlink missions.

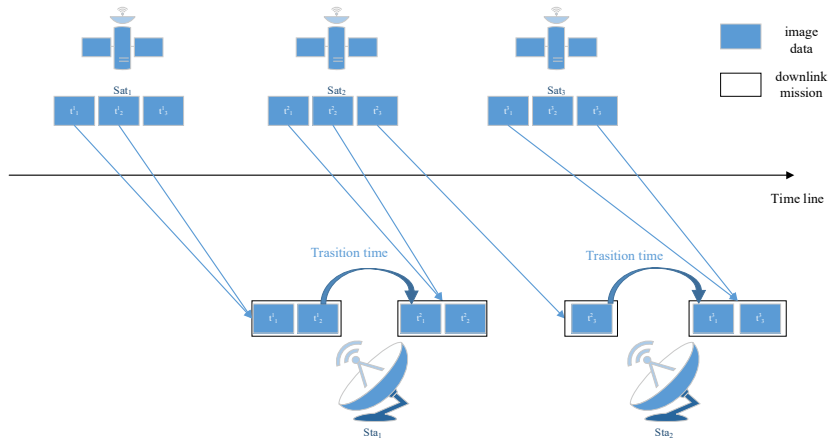

Figure 1 The process of SIDSP

However, the unequal development between the observation capability and the transition capability enriches the intension of SIDSP to a novel combinatorial problem, named as segmented satellite image data downlink scheduling problem (SSIDSP). It is similar to an one-dimensional two-stage cutting stock problem (TSCSP), as shown in Figure 2, in which the stock rolls should be cut into the intermediate rolls first, whose widths are not known a priori but restricted to lie within a specified interval, and in the second stage the finished rolls of demanded widths are produced from these intermediate rolls (Muter and Sezer, 2018).

In the first stage, three important issues should be considered to segment all OID. Whether an OID need to be segmented or not? How much SID should it be divided to? How much volume each SID should have? These three issues all are not fixed for every OID in advance. Note that, we would like to define this problem as OID cutting problem (OIDCP), which is similar to the first stage problem in the TSCSP.

In the second stage, we will consider SID and OID (it is not necessary to segment them) together to optimize the downlink scheme, the process of which is similar to the orthodox SIDSP. However, a novel characteristic named family attribute (FA) is a main distinguish between them. FA means all segmented image data from a same OID should be transmitted completely or not. Therefore, the second stage problem is named as satellite image data downlink scheduling problem with family attribute (SIDSPWFA).

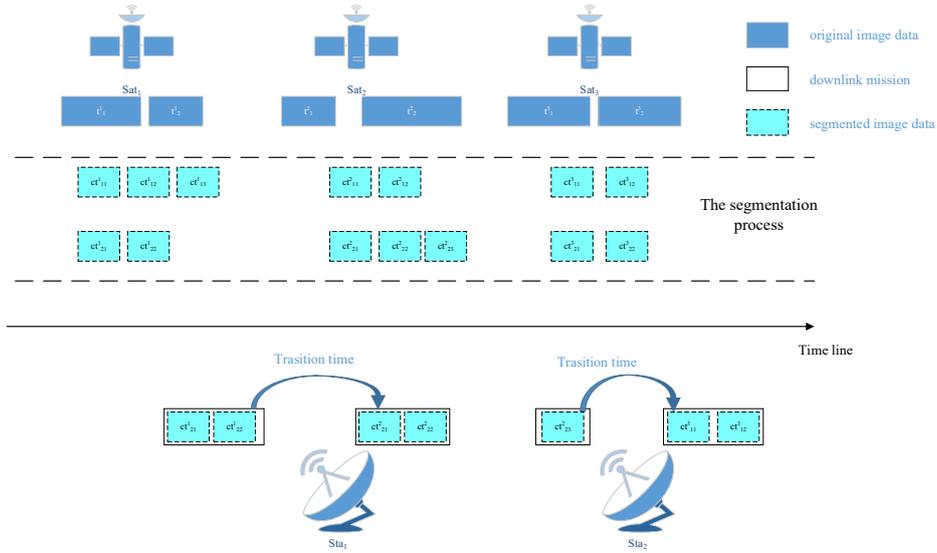

Figure 2 The process of SSIDSP

The paper is organized as follows. Section 2 develops a whole description of SIDSPWFA. In section 3 the mathematical formalization of SIDSPWFA is proposed. Section 4 designs a two-stage differential evolutionary algorithm (DE+NSGA-II) to solve SIDSPWFA. Extensive simulation instances and computational results are reported in section 5. Concluding remarks in the end.

Throughout this paper, we represent variable and parameter names using one or more operators following the convention of object-oriented programming which we believe will improve the readability and help ease of presentation of this complex problem. Note that these variables/parameters can alternatively be represented using one or more subscripts or superscripts. We preferred the object-oriented representation rather than the traditional mathematical representation of the constraints. For example, let $S$ be a parameter having attributes $x$ and $y$. Then, the x associated with $S$ is denoted by $S.x$ and y associated with $S$ is denoted by $S.y$. This notational convention is used throughout this paper without additional explanations.

## 2. The problem description

In this section, all preliminaries and mathematical descriptions are proposed. In addition, a fast segmentation operator is developed to segment OID in the minimum segmentation strategy.

## 2.1. Preliminaries

In order to better solve the focus in this paper, some reasonable assumptions are proposed to standardize and simplify SSIDSP according to previous researches and engineering experiences.

(1) We assume there is no any data loss of OID after segmenting. On the other hand, there is not any waiting time (conversion time) to receive different image data belongs to the same EOS to one ground station.
(2) Relay satellites actually can receive satellite image data, but they are only saved image data temporarily. That is, all image data saved in relay satellites have to be transmitted to ground stations again. So, we do not consider relay satellites in our study.
(3) Incompletely downlink is not considered our research. Note that, if only if an OID can be transmitted completely, it may be selected and added to the transmission scheme.
(4) We do not split a VTW to serval subs and then utilize them respectively. In other words, all VTWs can be used at most once.
(5) We assume there is only one antenna in every satellite and ground station. Therefore, one ground station can only receive image data from one satellite at most in any time. Or, a satellite can only transmit image data to one ground station at most in any time.
(6) Before transmitting image data, the transmission antenna in the satellite and the receiving antenna of the ground station should be calibrated, which is need waiting for some time named set-up time.
(7) It is apparent that the segmented image data much smaller, it is easier to full VTWs. But too small segmentation is bound to bring more difficulties to image processing in the ground and also not conducive to the satellite system operation. So, we do not consider over segmentation for image data and set a parament named minimum size of image data (MSID=10 s) to restrict segmentation.
(8) All existent OID and available VTWs are certain and static before downlink scheduling.

Based on the aforementioned assumptions, the scheme $\mathcal{S}$ for SSIDSP can be generally described by:
$$\mathcal{S} = \{S, G, St, Et, T, W, NT, M\} \quad (1)$$
where the included notations are defined as follows:
- $S = \{s | 1 \leq s \leq n_s, |S| = n_s\}$ is a set of all available earth observation satellites (EOSs) in SSIDSP.
$$s = \{Id, \Omega, i, a, e, \omega, M_0\} \quad (2)$$
where $Id$ is the identifier of $s$. $\{\Omega, i, a, e, \omega, M_0\}$ denotes the six orbital elements for $s$. As indicated earlier towards the end of the introduction section regarding notational convention, all attributes of $s$ will be represented in the format "$s.*$". For example, the identifier $Id$ of $s$ will be denoted by $s.Id$. Without further explanations, we follow this notational convention throughout this paper.

- $G = \{g | 1 \leq g \leq n_g, |G| = n_g\}$ is the set of all available ground stations for receiving image data in SSIDSP. Each $g$ is composed of a six-tuple.

$$g = \{Id, lat, lon, alt, \gamma, \pi\} \quad (3)$$

where $Id$ is the identifier of $g$. $\{lat, lon, alt\}$ denotes the location of $g$. $\gamma$ and $\pi$ denote the maximum angle of roll and pitch of the antenna installed in $g$. These two angles restrict the visibility of GSs with EOSs. in addition, before receiving image data, ground station will spend some time, named set-up time, to calibrate the antenna to EOS. Note that, we will define the set-up time following.

- $[St, Et]$ is the scheduling horizon for the SSIDSP. In our research, we fix the scheduling horizon as 24h from 2020/10/15 00:00:00 to 2020/10/16 00:00:00.
- $T = \{t_i | 1 \leq i \leq n_t, |T| = n_t\}$ represents the set of valid OID. $n_t$ indicates the quantity of OID. Each OID $t_i$ is composed of a seven-tuple.

$$t_i = \{Id, \omega, r, o, d, s, W_i\} \quad s \in S \quad (4)$$

where $Id$ is the identifier of $t_i$, $\omega$ reflects the priority of $t_i$, $r$ is the release time of $t_i$, $o$ is the due time of $t_i$, $d$ is the observation duration of $t_i$, $s$ is EOS that $t_i$ belongs to, and $W_i$ is the set of VTW can serve (receive) $t_i$, which will be defined as follows. Note that, the unit of the due time is hour. In addition, $[r, r + o]$ denotes the expiration date of $t_i$, which is used to confirm the set of available VTWs in $W_i$.

$$t_i.o = \begin{cases} 24 & t_i.\omega \in [1,3] \\ 12 & t_i.\omega \in [4,6] \\ 6 & t_i.\omega \in [7,9] \\ 3 & t_i.\omega = 10 \end{cases} \quad (5)$$

- $W = \{w_j | 0 \leq j \leq n_w, |W| = n_w\}$ indicates the set of all visible time windows (VTWs) during the scheduling horizon. Each $w_j$ has five important elements.

$$w_j = \{Id, sw, ew, s, g\} \quad s \in S, g \in G \quad (6)$$

where $Id$ is the identifier of $w_j$, $[sw, ew]$ is the available period of $w_j$, $g$ represents the ground station and $s$ is EOS that is visible with $g$ during $[sw, ew]$. Note that, all VTWs are sorted by their begin time $sw$ ascending.

- $W_i = \{w_j^i | 0 \leq j \leq n_w^i, |W_i| = n_w^i\}$ is the set of VTWs can serve (receive) $t_i$. Note that,

$w_j^i.s = t_i.s$, $w_j^i.sw \geq t_i.r$ and $w_j^i.sw < t_i.r + t_i.o$.

- $NT = \{nt_j | 0 \leq j \leq n_{nt}, |NT| = n_{nt}\}$ represents the set of all image data, including OID follows from $T$ without segmentation and segmented image data generated by OIDCP. Each image data $nt_j$ has seven attributes.

$$nt_j = \{Id, Fn, \omega, r, o, nd, s\} \quad (7)$$

where $\{Id, \omega, r, o, s\}$ inherit directly from the corresponding OID. $Fn$ is a positive integer variable and indicates the family identifier for $nt_j$, which is used to distinguish all segmented image data from a same OID. $nd$ toward $nt_i$ is similar to $d$ toward $t_i$ and indicates the downlink duration of $nt_i$.

- $M = \{m_j | 0 \leq j \leq n_m, |M| = n_m\}$ is the set of all downlink missions, which are

generated during the whole scheduling horizon. Each $m_j$ has six important elements.

$$m_j = \{Id, st, w, NT_j, s, g\} \qquad s \in S, g \in G \qquad (8)$$

where $Id$ is the identifier of $w_j$, which is the window for executing the mission $m_j$, $s$ and $g$ inherit directly from $w_j$. $st$ denotes the start time of $m_j$, and $w$ indicates the work duration time of $m_j$. $NT_j$ indicates the set of all image data transmitted in $m_j$. Note that $NT_j \subset NT$ and $m_j.w \geq \sum_{nt \in NT_j} nt.d \times rp$. Where $rp$ represents the recording and playback ratio, and we will set it in the simulation experiment.

The model and algorithm for OIDCP

In this section, we would like to first introduce two basic constraints, "Playback" and "Segmental", for segmentation. Then, using cutting stock problem (CSP) for reference, we will propose a minimum segmentation strategy to formalize OIDCP, and develop a fast segmentation operator.

## 2.2. A fast segmentation operator

To facilitate presentation, some notions are defined as below. Let an OID $t_i$ as an example.

- $NS$ indicates segmentation times.

$$NS = floor(\frac{t_i.d}{MSID}) \qquad (9)$$

where $floor(x)$ indicates the largest integer that does not exceed $x$.

- $RS$ indicates the remain duration after first segmentation.

$$RS = t_i.d - NS \times MSID \qquad (10)$$

The general process of segmentation operator can be described by the pseudo-code as Algorithm 1. All OID ($CT$), in the line 2:, that needs to be segmented can be filtered according to constraints (11) and constraints (12) easily. Note that, since the process to segment all OID is independent, every OID will be chosen randomly and segmented parallel.

| **Algorithm 1 :** A fast segmentation operator |
|---|
| **Input:** A set of OID $T$ |
| **Output:** A set of image data $NT$ including OID that does not need to be segmented, and segmented image data |
| 1:    Initialize the set $NT$ |
| 2:    Filter the set of OID need to be segmented ($CT$) according to the constraints (11) and constraints (12). While transform other OID to segmented image data and insert them into $NT$ directly. Note that, it is not necessary to segment them. |
| 3:    **While** $CT \neq \emptyset$ **do** |
| 4:       Choose an OID $t_i$ from $CT$ randomly. |
| 5:       Calculate the value of $NS$ and $RS$ toward $t_i$ using the function (9) and function (10). |
| 6:       Generate a set of segmented image data $NT^i = \{nt_j | 1 \leq j \leq NS, |NT| = NS\}$ based on the OID $t_i$. In addition, the downlink duration of $nt_j$ equals $MSID$. |
| 7:       Allocate $RS$ to each segmented image data averagely. |
| 8:       Insert $NT^i$ into $NT$ and delete $t_i$ from $CT$. |
| 9:    **End** while |

10:   Output all segmented image data.

    In addition, two constraints utilized in the Tab:1 are defined following respectively. As mentioned in the assumption 4, if only if an OID can be transmitted completely during the scheduling horizon, it is possible to consider it, otherwise it will be abandoned directly. This constraint is noted as Playback and defined as

$$t_i.d \le \sum_{w_j^i \in t_i.W_i}(ew - sw) \tag{11}$$

    No-over segmentation is a cardinal principle in our study. If only if an OID satisfies the Segmental constraint, it may be segmented into several SID. Let an OID $t_i$ as an example, this constraint can be described as

$$t_i.d > 2 \times MSID \tag{12}$$

where $MSID$ indicates the minimum size of image data (MSID).

## 3. The formalization of SIDSPWFA

    As mentioned above, SIDSPWFA is consist of two difficulty sub-problems, therefore, a fast segmentation operator is first developed to solve the first sub-problem, that is OIDCP. Considering SID and OID simultaneously, a bi-objective optimization model including several constraints is then built to format SIDSPWFA.

## 3.1. Two optimization objectives

    $NT = \{nt_i | 0 \le i \le n_{nt}, |NT| = n_{nt}\}$ indicates all image data need to be scheduled, including segmented image data (SID) and OID (OID). Decision variable $x_i^j$ marks whether image data $nt_i$ is transmitted in VTW $w_j$. That is,

$$x_i^j = \begin{cases} 1, if\ nt_i\ is\ transmitted\ in\ w_j \\ 0, otherwise \end{cases}$$

    Since the complexity of SIDSPWFA, the decision variable $x_i^j$ cannot represent all relationships. So, in order to facilitate to present the mathematical model, some notions and symbols are defined as below.

- $x_i = \begin{cases} 1, & if\ original\ image\ data\ t_i\ is\ scheduled \\ 0, & otherwise \end{cases}$
- $\bowtie$ is a computational operator, which indicates whether a downlink mission $m_j$ transmits any part of an OID $t_i$. Like $m_j \bowtie t_i \ne \emptyset$ means $m_j$ transmits some part of $t_i$. Otherwise, $m_j$ does not transmit any part of $t_i$.
- $N_{m_j \in M}(m_j \bowtie t_i \ne \emptyset)$ means the quantity of downlink missions which have transmitted some part of $t_i$, named as the segmentation times.

    It is an original intention of SIDSP to transmit as much OID as possible (Wang et al., 2011). Several optimization objective functions, maximize transmission revenue (Karapetyan et al., 2015, Marinelli et al., 2011), maximize downlink duration (Zhang et al., 2019) and minimize

transmission failure rate (Du et al., 2019), were adopted in many researches. Without loss of generality, we consider the image data transmission failure rate as one optimization objective in our study.

$f_1(S)$, ***the image data transmission failure rate (FR)***, namely, we consider both of the priority of image data and the downlink duration of image data in FR, which belongs to the interval [0,1].

$$f_1(S) = 1 - \frac{\sum_{i=1}^{n_t} x_i \times t_i.\omega \times t_i.d}{\sum_{i=1}^{n_t} t_i.\omega \times t_i.d} \tag{13}$$

On the other hand, optimizing the segmentation for every OID is another important goal in this paper. It is apparent that the segmentation times for each OID less is, the segmentation scheme better is. So, we design another optimization objective named ***the segmentation times (ST)***. Apparently, minimizing ST will decrease the quantity of used VTWs for every original data is transmitted. It is advisable to some extent.

$$f_2(S) = \frac{\sum_{i=1}^{n_t} x_i \times N_{m_j \in M}(m_j \bowtie t_i \neq \emptyset)}{n_t \times M_{st}} \tag{14}$$

where $x_i \times N_{m_j \in M}(m_j \bowtie t_i \neq \emptyset)$ represents ST of every scheduled OID. $n_t$ is the quantity of OID, and $M_{st}$ denotes the maximum segmentation times, defined as the function (15). By the way, $f_2(S)$ is normalized and distributed in the interval [0,1].

$$M_{st} = floor(\frac{\max_{t \in T} t.d}{MSID}) \tag{15}$$

where $MSID$ indicates the minimum size of image data (MSID). $\max_{t \in T} t.d$ denotes the maximum transmitted duration for all available OID. In addition, $floor(x)$ indicates a maximal integer that does not exceed $x$.

Given all that, we obtain two optimization objectives: ***the image data transmission failure rate (FR)*** and ***the segmentation times (ST)***. Moreover, these two objectives are not irreconcilable in satellite image data downlink scheduling, so the dual optimization of them is reasonable and possible.

$$min\ F(S) = \{f_1(S), f_2(S)\} \tag{16}$$

Observe that SIDSPWFA is a bi-objective discrete optimization problem, which has disconnected nondominated solution sets (Kidd et al., 2020). Then we would like to analyze all constraints in SIDSPWFA.

## 3.2. Main constraints describe

There are four main constraints in SIDSPWFA, including **Execution uniqueness constraint**, **The capacity constraint**, **The set-up time constraint** and **The completely transmitted constraint**, will formatted in detail.

### 3.2.1. Execution uniqueness constraint

As the name implies, the execution uniqueness constraint means every image data in the set $NT$ can be transmitted as most once. Let $\forall nt_i \in NT$ as an example and this constraint can be described as follows:

$$\sum_{j=0}^{|nt_i.W_i|} x_i^j \leq 1 \qquad (17)$$

where $nt_i.W_i$ denotes all VTWs can serve $nt_i$. In addition, $|nt_i.W_i|$ indicates the quantity of $t_i.W_i$.

### 3.2.2. The capacity constraint

Actually, the capacity of every VTW is finite. Let $\forall m_i \in M$ as an example and $w_j \in W$ is VTW according to $m_i$. This constraint can be described as follows:

$$\begin{cases} m_i.w \leq w_j.ew - w_j.sw \\ w_j.sw \leq m_i.st \\ m_i.st + m_i.w \leq w_j.ew \end{cases} \qquad (18)$$

The first function above indicates the work duration time of $m_i$ cannot exceed the length of $w_j$. Furthermore, combining the last two functions, it indicates $m_i$ must be executed during $w_j$.

### 3.2.3. The set-up time constraint

As mentioned in the assumption (5), one ground station only can receive image data from one satellite at most in any time. Note that, an interval time is used to turn the antenna from one satellite to another, named the set-up time ($\sigma_{s_i \to s_j}^g$), furthermore during $\sigma_{s_i \to s_j}^g$ no transmissions can occur (Marinelli et al., 2011). $\sigma_{s_i \to s_j}^g$ depend on rotation angle and rotating speed.

$$\sigma_{s_i \to s_j}^g = \frac{|Angle_{s_i} - Angle_{s_j}|}{rev_g} \qquad s_i, s_j \in S, g \in G \qquad (19)$$

where $Angle_{s_i}$ and $Angle_{s_j}$ denote angles when $g$ receives image data from $s_i$ and $s_j$ respectively.

Many studies (Du et al., 2019, Zhang et al., 2019, Zufferey et al., 2008) saw $\sigma_{s_i \to s_j}^g$ as a constant. While, Fabrizio (Marinelli et al., 2011) briefly discussed the value of $\sigma_{s_i \to s_j}^g$ under different work time of ground stations. They got a result that $\sigma_{s_i \to s_j}^g$ can be ignored when the work time of ground stations was bigger than 10 minutes. On the other hand, the set-up time is not our focus. So, we also would like to see $\sigma_{s_i \to s_j}^g$ as a constant and use $\sigma = 60\ s$ to denote it. Therefore, the set-up time constraint can be described as follows.

1. $\forall m_1, m_2 \in M$ transmits image dat from two different EOSs continuously, and both of them are executed in a same ground station, namely, $m_1.s \neq m_2.s$ && $m_1.g = m_2.g$
2. If $m_1.st > m_2.st$
3.     $m_2.st + m_2.w + \sigma \leq m_1.st$
4. else $m_1.st \leq m_2.st$
5.     $m_1.st + m_1.w + \sigma \leq m_2.st$
6. End if

### 3.2.4. The completely transmitted constraint

As mentioned in the assumption (3), if only if an OID can be transmitted completely, it may be scheduled to transmit. Otherwise, it should be abandoned. Let an OID $t_i$ as an example, this constraint can be described as follows:

$$\sum_{nt_j \in NT^i} \left( nt_j.nd \times \sum_{k=0}^{|W_i|} x_j^k \right) = x_i \times t_i.d \tag{20}$$

as defined above, $NT^i = \{nt_j\}$ indicates all segmented image data from an OID $t_i$.

## 4. A two-stage differential evolutionary algorithm

Recent decades witnessed the rapid development of multi-objective optimization methods centering evolutionary algorithms, and a large number of multi-objective evolutionary algorithms (MOEAs) has been proposed (Del Ser et al., 2019, Ramirez Atencia et al., 2019, Sun et al., 2019). Like NSGA (Srinivas and Deb, 1995), MOEA/D (Zhang and Li, 2007), MOEA/DD (Zheng et al., 2016), PICEA (Wang et al., 2013a), SPEA2(Adham et al., 2015) and so on. Note that, Deb (Deb et al., 2002) improved NSGA (Srinivas and Deb, 1995) to propose NSAG-II in 2002, which is one of the best evolutionary multi-objective optimization algorithms so far (Gong et al., 2009). NSGA-II obtained the Pareto frontier fast for a fast nondominated sorting approach, and solved the defect of the evolutionary algorithm depended on parameters by a fast-crowded distance estimation and a simple crowded comparison operator. Therefore, we would like to adopt NSGA-II as the evolutionary mechanism for our algorithm.

On the other hand, the differential evolution algorithm (DE) (Storn and Price, 1997) is a novel evolutionary algorithm for faster optimization and a simple yet powerful population based, direct search algorithm with the generation-and-test feature for globally optimizing functions using real valued parameters. Among the DE's advantages are its simple structure, ease of use, speed and robustness. Furthermore, it has been successfully used in solving some studies related to earth observation satellites (EOSs) (Chen et al., 2015, Li and Li, 2019, Yang et al., 2018). Therefore, we would like to adopt DE as the breeding algorithm for our algorithm.

| **Algorithm 2 :** A two-stage differential evolutionary algorithm (DE+NSGA-II) |
|---|
| **Input:** All original data *T*, all image data *NT* including OID and SID and all available VTW *W* |
| **Output:** The transmission scheme *M* |
| 1: SAll= **Initialize** (*NAll*, *NT*, *W*)    //Encode and generate the initial individuals |
| 2: **Improve** (SAll)          //Improve each individual by **insert operator** and **reorder operator** |
| 3: F_S = **Objective_function**( SAll )     //Obtain objectives value for each individual |
| 4: SBest = **Elitist_select** (SAll)          //Select the set of elitist individuals by NSGA-II |
| 5: **While Not termination** (Iteration) |
| 6:      SOffspring = **Generate** (SBest)     //Generate offspring individuals based on current elitist individuals by **mutation operator** and **swap operator** |
| 7:      **Improve** (SOffspring)          //Improve each individual by **insert operator** and **reorder operator** |
| 8:      SAll = **Combine** (SBest, SOffspring) //Update all individuals considering **box-method** |
| 9:      SBest = **Elitist_select** (SAll)     //Update the elitist individuals by NSGA-II |
| 10: End while |

11:    $M =$ **Decoding** (SBest)                   //Decode and obtain the transmission scheme

All in all, combing NSGA-II and DE, a two-stage differential evolutionary algorithm (DE+NSGA-II) is proposed to solve SIDSPWFA. The pseudo-code of which is described as Algorithm 2. Note that, the individual solution is two-stage, which we will describe in detail later, and the initial solutions are generated randomly in the line1:. Mind, the termination for DE+NSGA-II considers the maximum iteration time in the line 5:. Two types operators, in the line 6:, are designed to decrease the value of objectives and increase the diversity of solutions. Specifically, the first type operator, including insert operator and reorder operator, will decrease the value of two optimization objectives respectively, while the second type operator, including swap operator and mutation operator, is used to increase the diversity of solutions. In addition, a box-method (Hamacher et al., 2007) is utilized to decide whether new solutions are accepted or not in the line 8:.

## 4.1. Individual representation

As mentioned above, since SIDSPWFA should optimize segmentation scheme and the transmission scheme at the same time, the code of DE+NSGA-II is two-stage[1]. Encoding in the first stage represents whether an OID is scheduled and every code is a binary integer. While, the second stage code indicates the identification of VTW that is used to transmit image data.

To easy understand, a simple example is shown in Figure 3. In the first stage, $x_1 = 1$ denotes the OID, $t_1$, is scheduled. In addition, $x_1^2 = 1101$ and $x_1^3 = 1203$ convey two meanings. Firstly, $t_1$ is segmented into two segmented image data. Furthermore, those two segmented image data are transmitted in $1101^{st}$ VTW and $1203^{rd}$ VTW respectively.

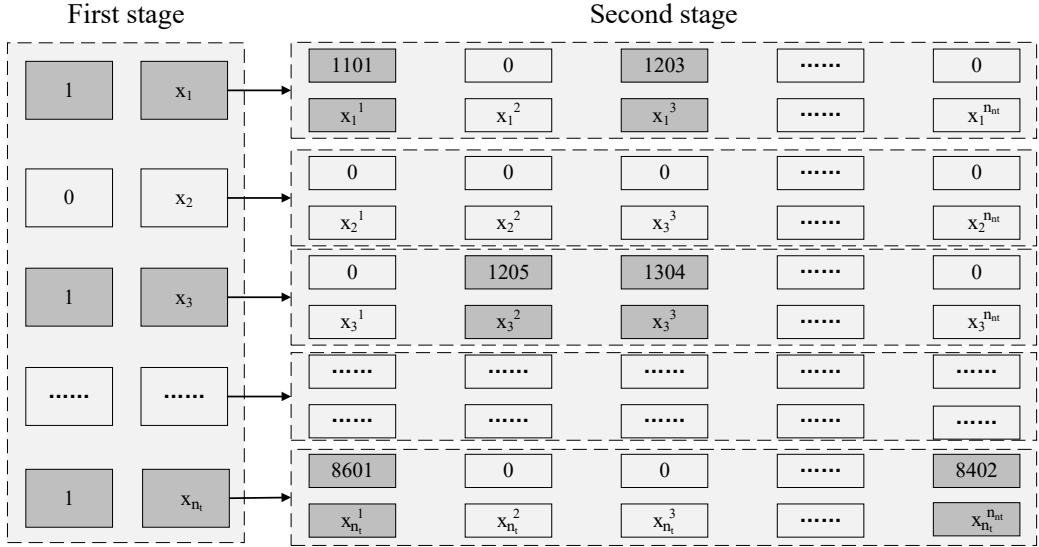

Figure 3 Individual representation

## 4.2. The bi-stage operators

A two-stage encoding results in bi-stage operators. The first stage operators, including

---

[1] At the beginning, all image data, including OID and SID, is sorted by their release time

insert operator and mutation operator, will be executed toward all OID. Insert operator improves the solution by decreasing FR, while mutation operator will increase the diversity of solutions by changing the transmitted scheme for OID. on the other hand, the second stage operators, including reorder operator and swap operator, are valid toward all segmented image data. Reorder operator can decrease ST of solutions, while swap operator is similar to mutation operator and also used to enrich the diversity of solutions. Note that, mutation operator and swap (cross) operator are two orthodox operators in the DE algorithm (Storn and Price, 1997), while insert operator and reorder are two heuristic operator, which are designed specially according to our study.

**4.2.1. The first stage operators**

As mentioned above, both two operators, insert operator and mutation operator, in the first stage will change the given transmission scheme facing OID directly, as shown in Figure 4. Note that, after the process of insert operator and mutation operator, a new solution is generated based on a given solution.

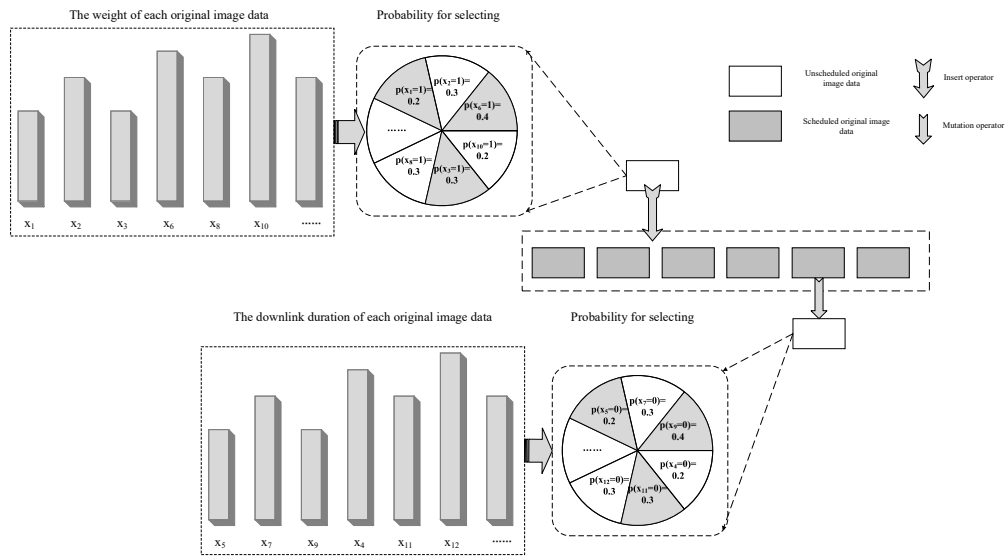

Figure 4 The two operators in the first stage

Insert operator will select some unscheduled OID by a probability, named as insert rate (IR), and then try to insert them into a given solution. Note that, we will discuss how to set the value of IR in section 5.3.1, which is critical for the insert operator.

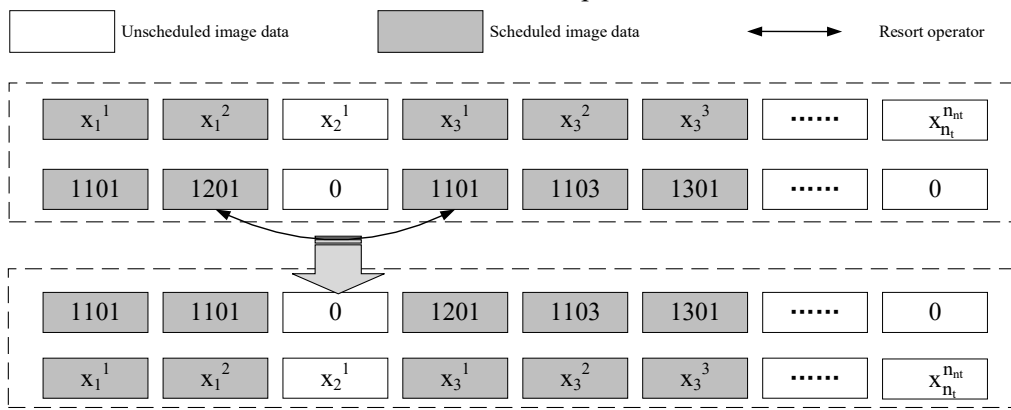

Figure 5 Reorder operator

While mutation operator will remove some scheduled OID from a given solution by a probability, named as mutation rate (MR), which we will also discuss in section 5.3.3, and by the way, mutation operator will generate more space for inserting other unscheduled OID.

**4.2.2. The second stage operators**

The reorder operator is used to improve every solution by decreasing their segmentation times (ST). Furthermore, the reorder operator may generate extra space for inserting other unscheduled OID. As shown in Figure 5. After reordering, two segmented image data $\{t_1^1, t_1^2\}$ from a same OID $t_1$ are transmitted in the same VTW ($Id = 1101$). It is apparent that ST of the given scheme is down by one at least.

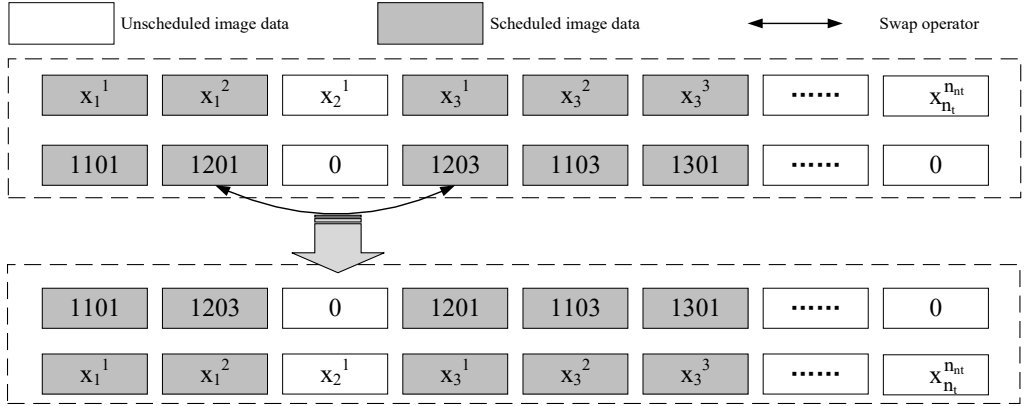

Figure 6 Swap inside of individual

While, the swap operator can be classified as two types according to the action objective. The first type swap operator is executed inside of a given solution as shown in Figure 6. Based on downlink duration of each segmented image data, it is easy to find any two segmented image data, like $x_1^2$ and $x_3^1$, can be exchanged and then generate a new solution. The other swap operator, as shown in Figure 7, exchanges some parts of parent-solutions randomly. If swapping succeeds, two new offspring solutions are generated.

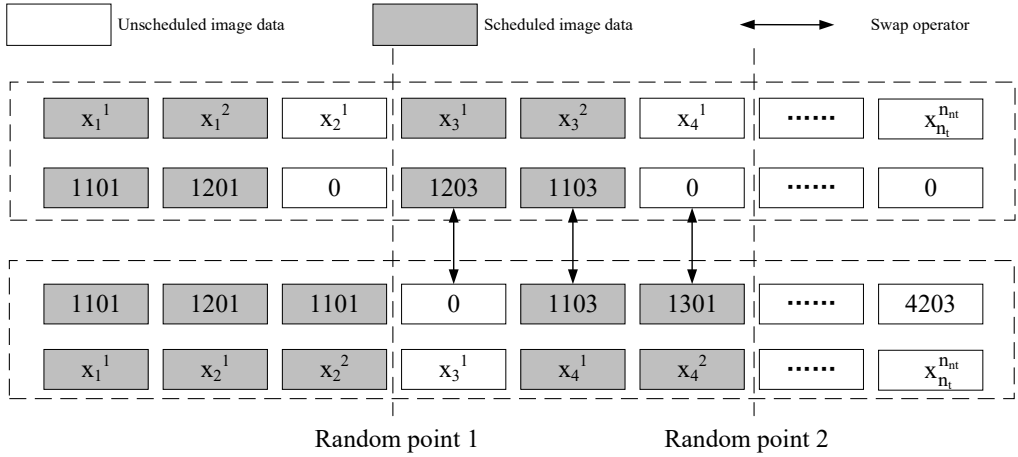

Figure 7 Swap among different individual

## 5. Simulation experiment

In this section, the influence of segmentation strategy on transmitting OID will be analyzed from several aspects in depth, moreover we would like to display the evolution and efficiency of DE+NSGA-II in advance. In addition, the extensive simulation instances for testing experiments are first designed according to the real world. Note that, we coded our algorithms in C#, using Visual Studio 2013, and performed our simulation experiments on a laptop with intel(R) Core (TM) i7-8750H CPU @ 2.2GHz and 16 GB RAM. Some general parameters in DE+NSGA-II are shown in Table 1.

Table 1 The parameters in DE+NSGA-II

| Parameter | Meaning | Value |
| --- | --- | --- |
| $NS$ | The population size of all preserving solution | 100 |
| $NA$ | The population size of archive solutions | 100 |
| $MaxIter$ | The maximum number of iterations | 50 |
| $IR$ | The possibility to control the insert operator | $0.4^2$ |
| $MR$ | The possibility to control the mutation operator and swap operator | 0.8 |

## 5.1. Design of simulation instances

Good test cases are necessary for simulation experiments. However, according to public researches, there is no any existent benchmark for SIDSP. Many researchers designed simulation instances according to the real world (Grasset-Bourdel et al., 2011, Karapetyan et al., 2015, Li et al., 2014, Xiao et al., 2019), without loss of generality, we would like to design simulation instances according the real world, considering three important aspects, that is available ground stations and earth observation satellites, and valid OID.

**5.1.1. Ground stations**

The first China remote sensing satellite ground station (RSGS) was formally established and operated in 1986 in Miyun (40°N/117°E), near Beijing. In recent years, Kashi Station (39°N/76°E), locates in Xinjiang Province in western China, and Sanya Station(18°N/109°E), locates in southern China's Hainan Province, have been constructed to expand satellite operations and reception coverage area (Guo et al., 2012). By the end of 2016, the China Remote Sensing Satellite North Polar Ground Station (CNPGS, 67°N/ 21°E), China's first overseas land satellite receiving station, was put into trial operation near Kiruna, Sweden (Na, 2016).

---

[2] If the possibility is bigger than the fixed value, $IR$ or $MR$, the corresponding operation will be done.

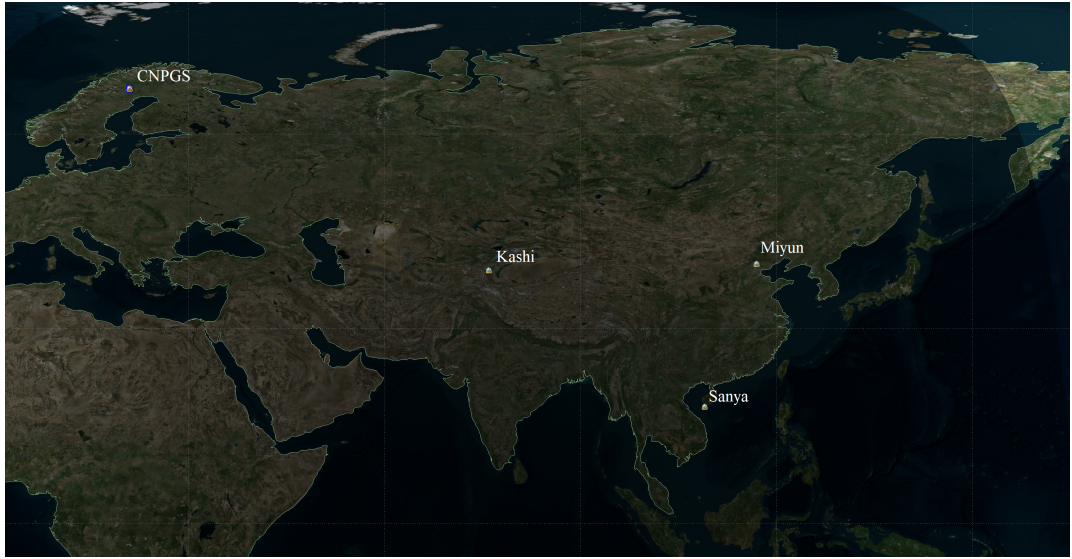

Figure 8 The location of available ground stations

Therefore, we would like to adopt these four ground stations, 3 normal ground station and 1 polar station, in our research. Moreover, the location of them is shown in Figure 8.

**5.1.2. Earth observation satellites**

We would like to choose ten state-of-the-art Chinese low orbit EOSs (LEOSs) from the satellite database in the AGI Systems Tool Kit (STK) 11.2, three EOSs are from the series of "Gao Fen" satellites, four EOSs are from the series of "Super View" satellites and other three of them are the series of "Earth Resource" satellites. In addition, the basic attributes of them are shown in Table 2. As mentioned above, the scheduling horizon as 24 hours, from 2020/10/15 00:00:00 to 2020/10/16 00:00:00. By the way, all transmission windows (TWs) of all available ground stations can be calculated easily by SKT. Note that, the whole data about orbits of EOSs, location of ground stations and available TWs are available on request from the first author. In addition, the time of all data is cumulative seconds based on 2020/10/15 00:00:00.

Table 2 The basic attributes of all available EOSs

| Name   | Id | $\Omega(km)$ | $i$   | $a$   | $e$    | $\omega$ | $M_0$  |
|--------|----|--------------|-------|-------|--------|----------|--------|
| GF0101 | 1  | 7145.08      | 0.001 | 98.55 | 359.06 | 152.17   | 265.39 |
| GF0201 | 2  | 7011.57      | 0.002 | 97.83 | 2.89   | 98.15    | 257.45 |
| GF0601 | 3  | 7020.45      | 0.002 | 97.99 | 6.87   | 56.94    | 94.33  |
| SV01   | 4  | 6901.65      | 0.002 | 97.43 | 1.01   | 124.24   | 242.68 |
| SV02   | 5  | 6894.39      | 0.001 | 97.54 | 11.87  | 128.22   | 90.39  |
| SV03   | 6  | 6883.14      | 0.000 | 97.51 | 5.98   | 341.26   | 106.70 |
| SV04   | 7  | 6884.95      | 0.004 | 97.51 | 6.14   | 92.52    | 195.65 |
| ZY02C  | 8  | 7143.90      | 0.002 | 98.64 | 341.91 | 57.55    | 186.17 |
| ZY3    | 9  | 6875.80      | 0.001 | 97.41 | 0.79   | 59.20    | 71.87  |
| ZY0104 | 10 | 7145.08      | 0.001 | 98.55 | 359.06 | 152.17   | 265.39 |

**5.1.3. Original image data**

Considering 3 normal ground stations and 1 polar station, there is about 1500 seconds downlink time (no considering overlap) for each EOS every day (about 14.5 orbits), so there is about 100 seconds during every orbit for each EOS to observe, which could be as a guidance for generating OID. We would like to design three types simulation instances according to adopted ground stations, Normal distribution (ND), Polar distribution (PD) and Mixed distribution (MD). ND only considers 3 normal ground stations, PD only adopts the unique polar ground station, while MD considers all available ground stations. Furthermore, the quantity of OID in every distribution is shown in Table 3.

Table 3 Three types realistic instances

| Distribution name | Available ground stations | Quantity of OID |
| --- | --- | --- |
| ND | Normal stations | [50, 500], with an increment step of 50 |
| PD | Polar station | [50, 500], with an increment step of 50 |
| MD | Normal stations + Polar station | [100,1000], with an increment step of 100 |

On the other hand, the priority of all OID is uniformly generated respectively from [1, 10], and the observation duration of them belongs to [10,200], with unit as second. Specifically, toward the three "Gao Fen" satellites, the observation duration belongs to [60,120]. Toward the four "Super View" satellites, the observation duration belongs to [10,60], while for other "Earth Resource" satellites, the observation duration belongs to [120,200]. Note that, we assume these ten satellites have the same imaging capability and transmission capability as the four "Super View" satellites, that is 2×450Mbps (Sai et al., 2018), and recording and playback ratio of them all equals 1:4, that is $rp = 4$ ($rp$ of GF-2 is 1:4.8 (Huang et al., 2018)), which means it will spend 4 minutes to transmit the image data formed by one-minute observation.

In addition, the release time of all OID is uniformly distributed in the time interval from 2020/10/14 00:00:00 to 2020/10/16 00:00:00, and the due time of them is calculated by the function (5). According to the release time and the due time, we can filter all valid OID in the scheduling horizon (2020/10/15 00:00:00 to 2020/10/16 00:00:00). Note that, all OID is available on request from the first author.

## 5.2. The convergence and efficiency of DE+NSGA-II

Since some evidence exists show that the random search can be competitive to evolutionary approaches in multi-objective spaces (Corne and Knowles, 2007, Purshouse and Fleming, 2007, Wang et al., 2013b), we propose a very crude random evolutionary mechanism (CREM), in which the elitist solutions are preserved randomly, to analyze the evolutionary mechanism, NSGA-II, used in our algorithm. Combining then CREM with the DE algorithm, note as DE+CREM, as the control algorithm.

All simulation instances in the Polar distribution (PD) are chosen as test cases and the value of Hypervolume (HV) is considered as the critical indicator to analyze the convergence of our algorithm, DE+NSGA-II, in detail. Note that, HV is calculated by an influential algorithm, Hypervolume by slicing objectives (HSO) (Bradstreet et al., 2008, Durillo and Nebro, 2011), which is an exact calculation method with high accuracy. The iteration values of HV obtained by these two algorithms, DE+NSGA-II and DE+CREM, respectively are shown in Figure 9, where ten sub plots correspond to the ten test cases respectively, and the black dotted line and the blue dotted line denote the iteration values of HV obtained by DE+NSGA-II and DE+CREM respectively.

It is apparent that the iteration values of HV obtained by DE+NSGA-II always reaches a stable value or trend after about 50 iterations for all test cases, which is a good guidance information to adopt DE+NSGA-II for other simulation experiments. While the iteration values of HV obtained by DE+CREM are always in a disordered state. Therefore, the convergence of DE+NSGA-II is confirmed to some extent.

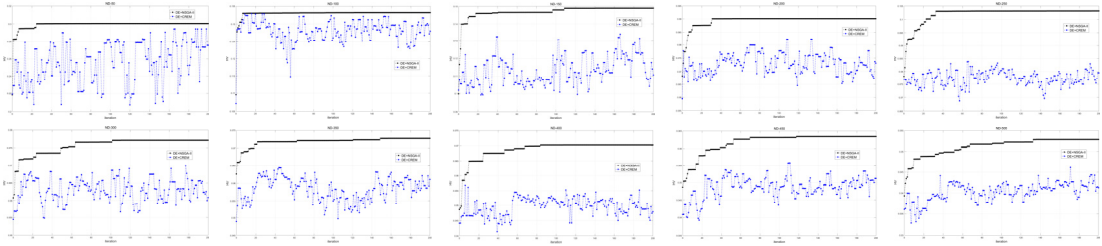

Figure 9 The iteration values of HV during 200 obtained by DE+NSGA-II and DE+CREM

In order to explain the convergence of DE+NSGA-II, we set $MaxIter = 50$ and restart these two algorithms for 50 times respectively based on all test case. Furthermore, a boxplot of the average HV obtained by them as shown in Figure 10. In addition, in order to display the efficiency of DE+NSGA-II more visually, the Pareto frontier of five specific test cases are also drawn, PD-100~PD-500, with an increasement step of 100. Note that, they are obtained under the best value of HV during 50 times restarting.

Firstly, the length of black boxes is significantly shorter than that of blue boxes, and the quantity of red pluses for black boxes is always fewer than that for the blue boxes, both of which reflect DE+NSGA-II can achieve the nondominated solutions stably, which is fully supports the previous conclusion, that is the convergence of DE+NSGA-II is very well.

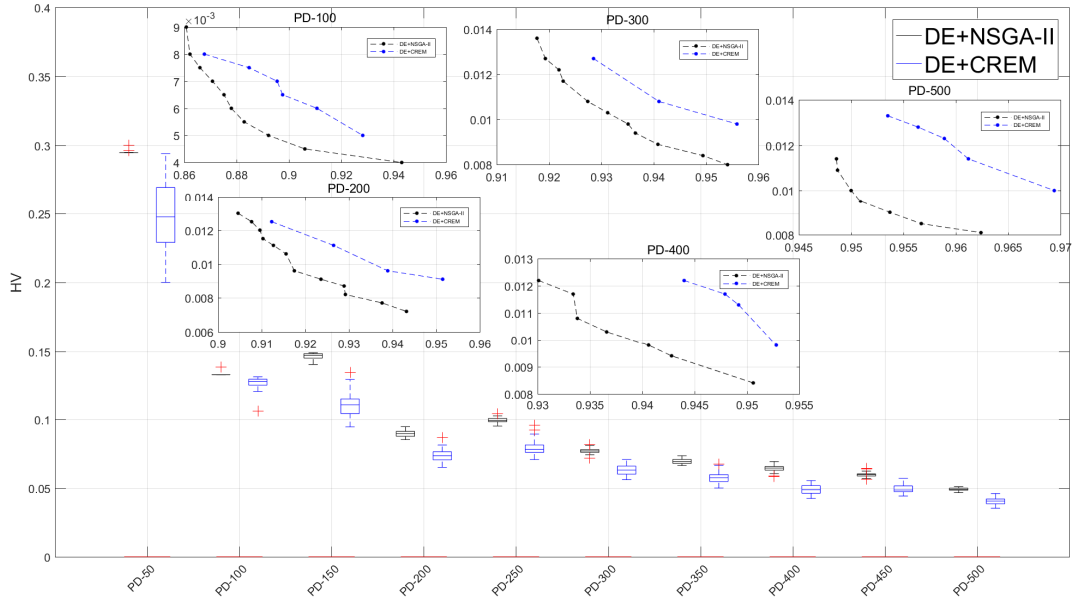

Figure 10 Boxplot: the restarting values of HV obtained by DE+NSGA-II and DE+CREM, and Dotted

line: the best Pareto frontiers of five specific test cases

Secondly, it reflects the nondominated solutions obtained by DE+NSGA-II are constantly and significantly better than that obtained by DE+CSEM that the location of black boxes is always higher than that of blue boxes for all simulation instances.

Thirdly, toward five specific test cases, the location of Pareto frontiers also reflects that DE+NSGA-II has a nice efficiency. On the one hand, the location of Pareto frontiers obtained by DE+NSGA-II is always under that obtained by DE+CSEM. On the other hand, the Pareto frontiers obtained by DE+NSGA-II is also constantly and significantly longer, longer Pareto frontier is, more diverse solutions are.

All in all, DE+NSGA-II is not only with a nice convergence, but also good at searching much better and more diverse nondominated solutions.

## 5.3. The evolution of all proposed operators

We first would like to the evolution and efficiency of these two-stage operators, including two orthodox operators, mutation operator and swap operator, and two heuristic operators, insert operator and reorder operator. Furthermore, based on extensive simulation results, the suitable value of some critical paraments will be confirmed. Note that, according to the convergence of DE+NSGA-II above, we would like to set $MaxIter = 50$ for all simulation experiments.

### 5.3.1. Insert operator

A simulation instance, ND-300 in the Normal distribution, is chosen to analyze the evolution of insert operator. Let $IR$ belongs to the interval [0,1] steps 0.1, and adopt DE+NSGA-II for under different values of $IR$ based on ND-300. The final value of HV and the value of the bi-objective obtained by DE+NSGA-II for under different values of $IR$ are shown in Table 4. Where $\widehat{FR}$, $\overline{FR}$ and $\widetilde{FR}$ denote the maximum, average and minimum value of FR, while $\widehat{ST}$, $\overline{ST}$ and $\widetilde{ST}$ denote the maximum, average and minimum value of ST.

Table 4 The final value of HV and the value of the bi-objective

| $IR$ | HV | $\widehat{FR}$ | $\overline{FR}$ | $\widetilde{FR}$ | $\widehat{ST}$ | $\overline{ST}$ | $\widetilde{ST}$ |
|---|---|---|---|---|---|---|---|
| 0 | 0.0947 | 0.95 | 0.9156 | 0.9044 | 0.0127 | 0.0108 | 0.009 |
| 0.1 | 0.1055 | 0.9491 | 0.9069 | 0.8935 | 0.0134 | 0.0112 | 0.009 |
| 0.2 | 0.1081 | 0.9468 | 0.9056 | 0.8909 | 0.0123 | 0.0104 | 0.009 |
| 0.3 | 0.1077 | 0.9298 | 0.9047 | 0.8913 | 0.0127 | 0.0108 | 0.0087 |
| **0.4** | **0.1093** | **0.9313** | **0.9023** | **0.8897** | **0.0134** | **0.011** | **0.009** |
| 0.5 | 0.1073 | 0.9345 | 0.9017 | 0.8918 | 0.013 | 0.0103 | 0.0083 |
| 0.6 | 0.1054 | 0.9377 | 0.9091 | 0.8937 | 0.0112 | 0.0094 | 0.008 |
| 0.7 | 0.1056 | 0.9337 | 0.9041 | 0.8936 | 0.0108 | 0.0091 | 0.0069 |
| 0.8 | 0.1068 | 0.9351 | 0.9041 | 0.8925 | 0.0108 | 0.0085 | 0.0058 |
| 0.9 | 0.1046 | 0.9687 | 0.9189 | 0.8949 | 0.0112 | 0.0071 | 0.0029 |
| 1 | 0.0854 | 0.9934 | 0.9402 | 0.9143 | 0.0098 | 0.0049 | 0.0004 |

Since the final value of HV receives its maximum (0.1093) when the value of $IR$ equals 0.4, it may be more suitable to some extent setting the value of $IR$ as 0.4 to adopt DE+NSGA-II. But actually, the final values of HV obtained by DE+NSGA-II under different values of $IR$ are not different significantly, except $IR$ equals 0 and 1. When the value of $IR$ equals 0, insert operator will be done every time, while when the value of $IR$ equals 1, insert operator is invalid. Under both of these two conditions, the final values of HV receive the worse values than other conditions. Therefore, we would not consider them.

In addition, in order to show the difference of HV and bi-objective obtained by DE+NSGA-II under different values of $IR$ visually, the Pareto frontier and the iteration values of HV obtained by DE+NSGA-II under all conditions after 50 iterations are drawn together in Figure 11. The above plot corresponds to the Pareto frontier, while the other one represents the iteration values of HV. In addition, all labels are also marked apparently.

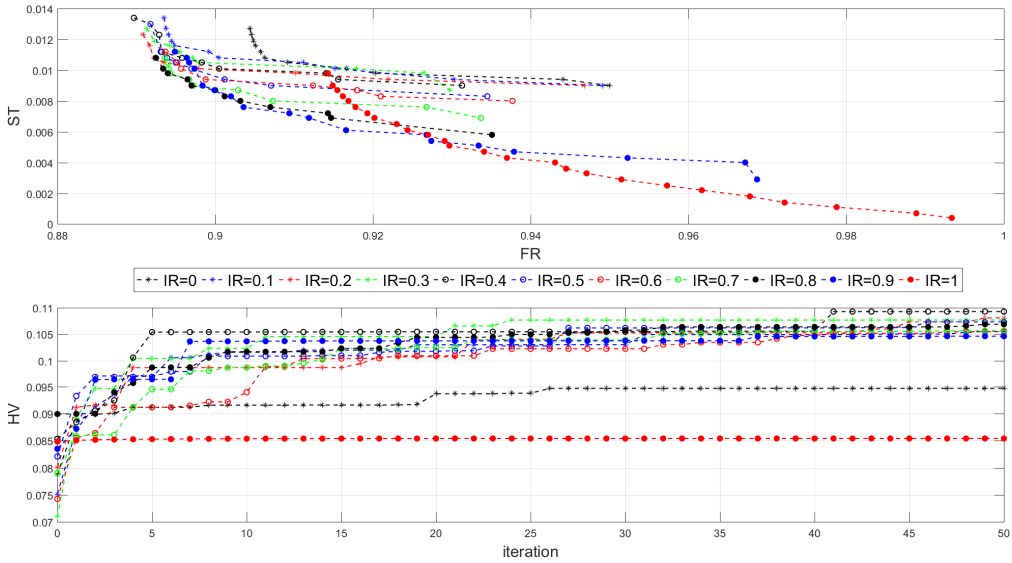

Figure 11 The Pareto frontier and the iteration values of HV obtained by DE+NSGA-II under different

values of *IR*

Firstly, it reflects both the two values, 0 and 1, are the worse values further from the location of Pareto frontier and iteration values of HV obtained by DE+NSGA-II under *IR* equals 1 and 0. That is, the Pareto frontier under them are dominated by that under other conditions, and the iteration values of HV under them are less than that under other conditions.

Secondly, the efficiency of insert operator are similar under *IR* belongs to the interval [0.1,0.9] actually, therefore, it is no any significant evidence to prefer to some specifical values.

All in all, setting *IR* as 0.4 is a relatively good choice for adopting DE+NSGA-II for simulation experiments.

**5.3.2. Reorder operator**

As mentioned above, the reorder operator is a heuristic operator and is designed to decrease the value of ST directly, meanwhile it may generate extra space for inserting other unscheduled OID, by the way, the value of FR may also be decreased. To analyze the efficiency and evolution of reorder operator, all simulation instances in the Normal distribution are considered as test cases, and then we will adopt DE+NSGA-II with reorder operator and without reorder operator respectively based on all test cases. Note that, the same initial solutions are adopted for them toward each case respectively. The Pareto frontier and the iteration values of HV for every case are drawn together in Figure 12. The ten subplots correspond to the ten cases respectively. The main figure denotes the Pareto frontier, while the secondary one represents the iteration values of HV. In addition, all designations are also marked significantly.

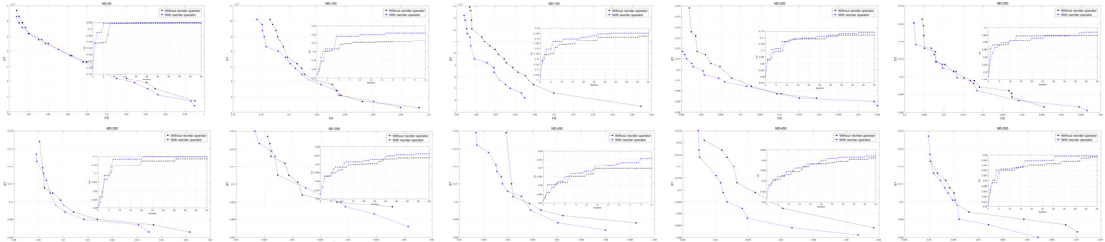

Figure 12 The Pareto frontier and the iteration values of HV for all test case obtained by DE+NSGA-II

An important and apparent observation is that the iteration values of HV obtained by DE+NSGA-II considering reorder operator constantly and always outperform that without considering reorder operator. In other words, the quality of solutions searched by DE+NSGA-II is improved more significantly when considering reorder operator.

It also reveals the same result from the location of Pareto frontier obtained by DE+NSGA-II with the reorder operator or not respectively. The Pareto frontier obtained by DE+NSGA-II with the reorder operator (the blue dotted line) is always under that without the reorder operator (the black dotted line).

### 5.3.3. Mutation operator and swap operator

As mentioned above, a same parameter, $MR$, is used to control mutation operator and swap operator, because the mechanism of them are similar. Note that, mutation operator works for OID, while swap operator faces segmented image data directly. In order to analyze the efficiency and evolution of them, we also choose the simulation instance, ND-300 in the Normal distribution, as an example. Let $MR$ belongs to the interval [0,1] steps 0.1, and restart DE+NSGA-II for under different values of $MR$ based on ND-300. The final value of HV and the value of the bi-objective obtained by DE+NSGA-II for under different values of $MR$ are shown in Table 5. Where $\widehat{FR}$, $\overline{FR}$ and $\widetilde{FR}$ denote the maximum, average and minimum value of FR, while $\widehat{ST}$, $\overline{ST}$ and $\widetilde{ST}$ denote the maximum, average and minimum value of ST.

Since the final value of HV receives its maximum (0.1082), under the value of $MR$ equals 0.8, it may be more suitable to some extent setting the value of $MR$ as 0.8 to adopt DE+NSGA-II. But actually, the final values of HV obtained by DE+NSGA-II under different values of $MR$ are not different significantly, especially toward to the values of 0.8 and 0.9. But when the value of $MR$ equals 1, which means mutation operator and swap operator are invalid, the final value of HV is significantly less than that under all other conditions. It reflects mutation operator and swap operator may be necessary to DE+NSGA-II to some extent.

Table 5 The final value of HV and the value of the bi-objective

| IR | HV | $\widehat{FR}$ | $\overline{FR}$ | $\widetilde{FR}$ | $\widehat{ST}$ | $\overline{ST}$ | $\widetilde{ST}$ |
| --- | --- | --- | --- | --- | --- | --- | --- |
| 0 | 0.0963 | 0.9328 | 0.9142 | 0.9029 | 0.0108 | 0.0097 | 0.008 |
| 0.1 | 0.0981 | 0.9406 | 0.9136 | 0.901 | 0.013 | 0.0103 | 0.0083 |
| 0.2 | 0.0946 | 0.9463 | 0.9194 | 0.9046 | 0.0119 | 0.0095 | 0.0076 |
| 0.3 | 0.0991 | 0.9185 | 0.9069 | 0.9 | 0.0123 | 0.0103 | 0.0083 |
| 0.4 | 0.1000 | 0.9561 | 0.92 | 0.8991 | 0.0116 | 0.0094 | 0.008 |
| 0.5 | 0.0986 | 0.9542 | 0.9155 | 0.9006 | 0.0127 | 0.01 | 0.008 |
| 0.6 | 0.1055 | 0.9305 | 0.9059 | 0.8936 | 0.0123 | 0.0103 | 0.0083 |
| 0.7 | 0.1037 | 0.9511 | 0.916 | 0.8954 | 0.0134 | 0.0096 | 0.0072 |
| **0.8** | **0.1082** | **0.9114** | **0.8995** | **0.8909** | **0.0123** | **0.0103** | **0.008** |
| **0.9** | **0.1080** | **0.9066** | **0.8957** | **0.8911** | **0.0123** | **0.0105** | **0.0083** |
| 1 | 0.0869 | 0.9633 | 0.927 | 0.9123 | 0.0123 | 0.0104 | 0.0083 |

In addition, in order to show the difference of HV and bi-objective obtained by DE+NSGA-II under different values of $MR$ visually, the Pareto frontier and the iteration values of HV obtained by DE+NSGA-II under all conditions after 50 iterations are drawn together in Figure 11. The above plot corresponds to the Pareto frontier, while the other one is used to draw the iteration values of HV. In addition, all labels are also marked apparently.

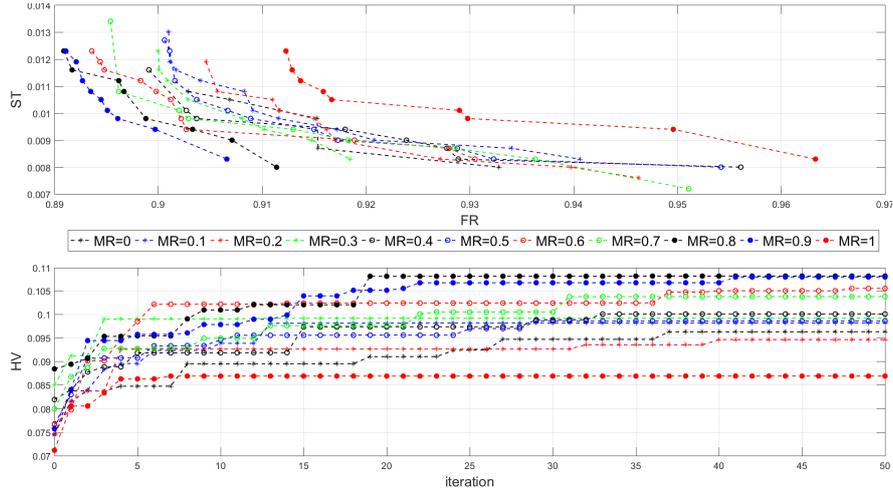

Figure 13 The Pareto frontier and the iteration values of HV obtained by DE+NSGA-II under different values of $MR$

Firstly, the iteration values of HV obtained by DE+NSGA-II under different values of $MR$ can be classified three levels. The iteration values of HV under the value of $MR$ belongs to the interval [0.6,0.9] are named as the first level, the iteration values of HV under the value of $MR$ belongs to the interval [0,0.5] are the second level, while the iteration values of HV under the value of $MR$ equals 1 are the third level. During the same level, the iteration values of HV are close. In addition, during the first level, mutation operator and swap operator are always more effect to some extent.

Secondly, it follows from the location of the Pareto frontier under different values of $MR$ that DE+NSGA-II under the value of $MR$ equals 0.9 slightly outperforms that under other conditions. Observe that the Pareto frontier obtained by DE+NSGA-II under the value of $MR$ equals 0.9 is located under that under other conditions.

All in all, actually speaking, the efficiency of mutation operator and swap operator are similar. However, considering all results together, $MR = 0.9$ is more suitable to adopt DE+NSGA-II for all simulation experiments.

## 5.4. The efficiency of segmentation

Two segmentation strategies, the no-segmentation strategy and stochastic segmentation strategy, are proposed as two control groups to discuss the efficiency of our segmentation strategy, the minimum segmentation. The no-segmentation strategy denotes that all OID will be transmitted directly without segmentation. If an OID cannot be transmitted completely in any VTW, it will be abandoned. While under the stochastic segmentation strategy, all OID will be segmented randomly and each segmented image data must satisfy the minimum segmentation constraint, the function (12).

All simulation instances in the Mixed distribution are chosen as test cases, and DE+NSGA-II is run under these three segmentation strategies respectively. Firstly, the iteration values of HV and the Pareto frontier are used as two evaluation indicators. The compared results are shown in Figure 14. Among them, the main plot displays the Pareto frontier, while the subplot corresponds the iteration values of HV. In addition, the black dotted line, the blue dotted line and the red dotted line represent the results obtained by DE+NSGA-II under these segmentation strategies respectively. Furthermore, the corresponding relationship is marked significantly in the figure.

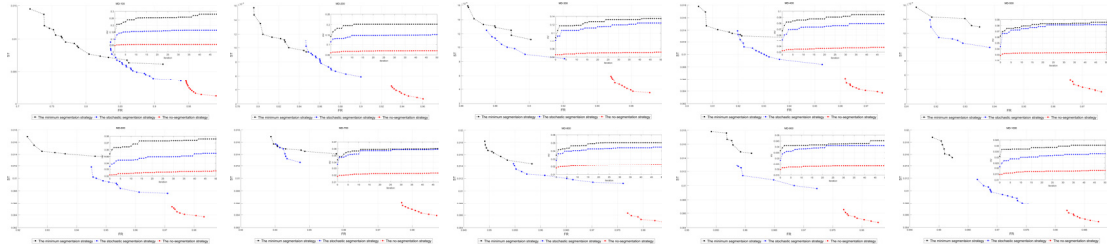

Figure 14 The Pareto frontier and the iteration values of HV obtained by DE+NSGA-II under different

segmentation strategies

It follows from the iteration values of HV obtained by DE+NSGA-II under different segmentation strategies that the minimum segmentation strategy outperforms other two segmentation strategies no matter what the scale of simulation instance is.

However, the Pareto frontier obtained by DE+NSGA-II under the minimum segmentation strategy cannot always dominate that under other two segmentation strategies. Some elitist solutions obtained by DE+NSGA-II the stochastic segmentation strategy even better than a part of the elitist solutions obtained by DE+NSGA-II under the minimum segmentation strategy, like MD-300 and MD-400. Actually speaking, the value of ST is always higher for more OID transmitted successfully, lower value of FR, under the minimum segmentation strategy. Therefore, the distribution of the Pareto frontiers obtained by DE+NSGA-II under these segmentation strategies is reasonable.

As mentioned in section 5.1.3, the observation duration varies according to the type EOS, "Earth Resource" satellites take the longest observation duration, while "Super View" satellites get the shortest. Therefore, in order to reveal the advantage of the minimum segmentation strategy, another evaluation indicator is added, named as the single success rate (SSR), in which the ratio of transmitted OID is calculated corresponding to each type EOS, as drawn by four subplots in Figure 15. The first subplot denotes the finial value of HV obtained by DE+NSGA-II under different segmentation strategies. While the other three subplots represent the single success rate under different segmentation strategies respectively. Note that, the maximum value and minimum value of SSR on the Pareto frontier are drawn together. In addition, all designations are also marked significantly.

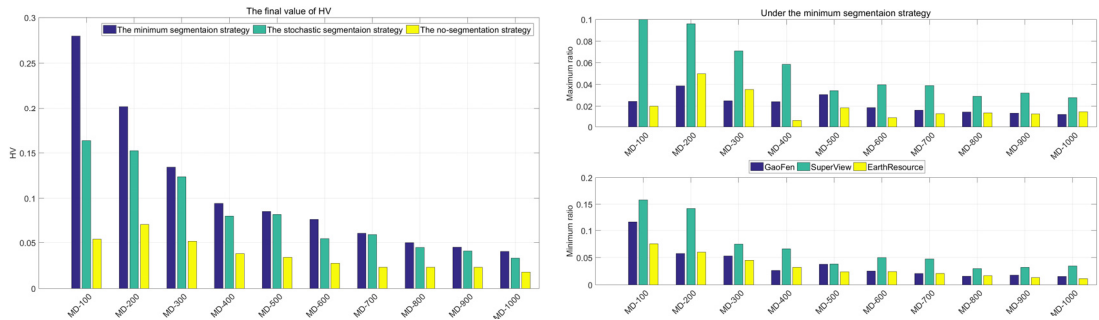

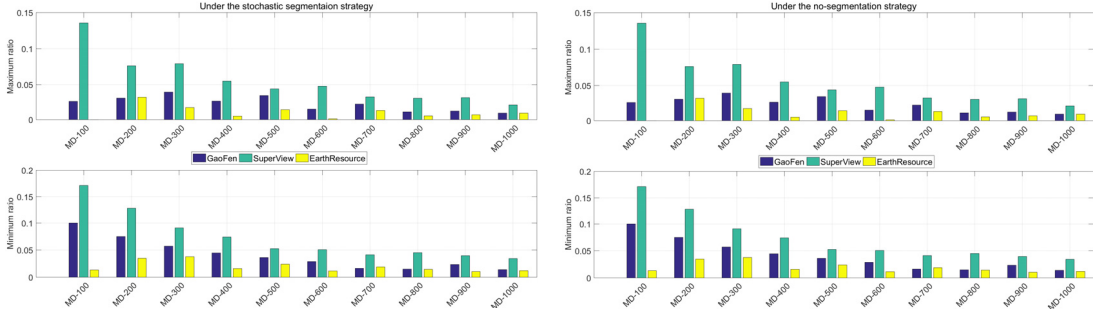

Figure 15 The final value of HV and the single success rate obtained by DE+NSGA-II under different

segmentation strategies

It follows from the final value of HV obtained by DE+NSGA-II under different segmentation strategies that the minimum segmentation strategy clearly outperforms other two segmentation strategies. In addition, observe that the final values of HV under segmentation, no matter the minimum segmentation strategy or the stochastic segmentation strategy, are all significantly better than that under the no-segmentation strategy.

Since the value of SSR for "Super view" satellites is always the biggest no matter under what segmentation strategies, OID belongs to "Super view" satellites is always and significantly transmitted is more than other two types EOSs. As mentioned above, observation duration of "Super view" satellites is the shortest, and some of them even can be transmitted completely without segmentation.

It is apparent that under the minimum segmentation strategy the value of SSR for different types EOSs is more well-distributed. Toward some specific test cases, MD-100 and MD-600, the extreme phenomena even arise that no any OID belongs to the "Earth Resource" satellites is transmitted successfully under the no-segmentation strategy and the stochastic segmentation strategy.

All in all, the minimum segmentation strategy is not only good for transmitting more and higher OID, but also conducive to balance SSR of different types EOSs. Furthermore, the efficiency of the minimum segmentation strategy will be more significant with the larger gap between the observation capability and the transition capability.

## 6. Conclusion

A novel combinatorial optimization problem, satellite image data downlink scheduling problem with family attribute (SIDSPWFA), was studied in our paper. Two optimization objectives, FR and ST, were designed to formalize SIDSPWFA as a bi-objective discrete optimization model. Furthermore, considering NSGA-II and DE, a two-stage differential evolutionary algorithm (DE+NSGA-II) was developed to solve SIDSPWFA.

The extensive simulation instances based on the real world display DE+NSGA-II is not only good at convergence, but also conducive to search more better and diverse non-dominated solutions.

SIDSPWFA is similar to a one-dimensional two-stage cutting stock problem (TSCSP) and a discrete bi-objective optimization problem, so the algorithms and operators proposed in our paper may be used to solve other similar problems in such a way.

# Acknowledgements

This work is supported by the science and technology innovation Program of Hunan Province (2021RC2048) of Zhongxiang Chang.